\documentclass[10pt,twocolumn,letterpaper]{article}

\usepackage{cvpr}
\usepackage{times}
\usepackage{epsfig}
\usepackage{graphicx}
\usepackage{amsmath}
\usepackage{amssymb}
\usepackage{multirow}
\usepackage{bm}
\usepackage{subcaption}
\usepackage[pagebackref=true,breaklinks=true,letterpaper=true,colorlinks,bookmarks=false]{hyperref}

\cvprfinalcopy 

\def\cvprPaperID{} 

\begin{document}

\title{Harvesting Multiple Views for Marker-less 3D Human Pose Annotations}

\author{Georgios Pavlakos$^1$, Xiaowei Zhou$^1$, Konstantinos G. Derpanis$^2$, Kostas Daniilidis$^1$ \\[0ex]
$^1$ University of Pennsylvania \hspace{2em} $^2$ Ryerson University\\ 
}

\maketitle

\begin{abstract}

Recent advances with Convolutional Networks (ConvNets) have shifted the bottleneck for many computer vision tasks to annotated data collection.  In this paper, we present a geometry-driven approach to automatically collect annotations for human pose prediction tasks. Starting from a generic ConvNet for 2D human pose, and assuming a multi-view setup, we describe an automatic way to collect accurate 3D human pose annotations. We capitalize on constraints offered by the 3D geometry of the camera setup and the 3D structure of the human body to probabilistically combine per view 2D ConvNet predictions into a globally optimal 3D pose. This 3D pose is used as the basis for harvesting annotations. The benefit of the annotations produced automatically with our approach is demonstrated in two challenging settings: (i) fine-tuning a generic ConvNet-based 2D pose predictor to capture the discriminative aspects of a subject's appearance (i.e.,``personalization"), and (ii) training a ConvNet from scratch for single view 3D human pose prediction without leveraging 3D pose groundtruth. The proposed multi-view pose estimator achieves state-of-the-art results on standard benchmarks, demonstrating the effectiveness of our method in exploiting the available multi-view information.

\end{abstract}

\section{Introduction}

Key to much of the success with Convolutional Networks (ConvNets) is the availability of abundant labeled training data. For many tasks though this assumption is unrealistic. As a result, many recent works have explored alternative training schemes, such as unsupervised training \cite{garg2016unsupervised,long2016learning,yu2016back}, auxiliary tasks that improve learning representations~\cite{wu2016single}, and tasks where groundtruth comes for free, or is very easy to acquire~\cite{pinto2016curious}. Inspired by these works, this paper proposes a geometry-driven approach to automatically gather a high-quality set of annotations for human pose estimation tasks, both in 2D and 3D.

\begin{figure}[t]
	  \centering
	  \includegraphics[width=0.99\linewidth,trim={4.5cm 5.5cm 4.5cm 6cm},clip]{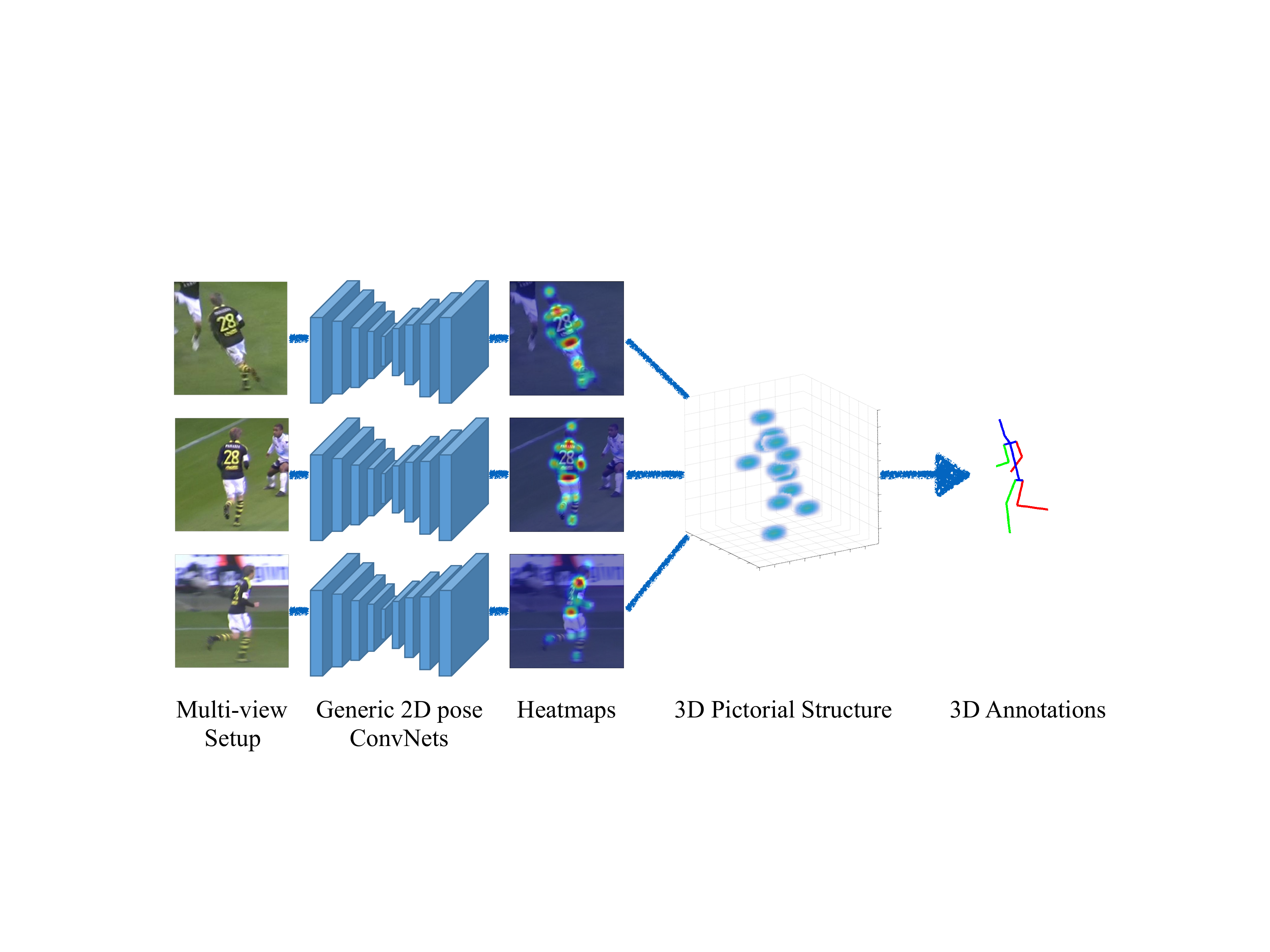}
    \caption{
Overview of our approach for harvesting pose annotations. Given a multi-view camera setup, we use a generic ConvNet for 2D human pose estimation~\cite{newell2016stacked}, and produce single-view pose predictions in the form of 2D heatmaps for each view. The single-view predictions are combined optimally using a 3D Pictorial Structures model to yield 3D pose estimates with associated per joint uncertainties. The pose estimate is further probed to determine reliable joints to be used as annotations.
}
 \label{fig:overview1}
\end{figure}

ConvNets have had a tremendous impact on the task of 2D human pose estimation \cite{toshev2014deep,wei2016convolutional,newell2016stacked}. A promising research direction to improve performance is to automatically adapt (i.e., ``personalize'') a pretrained ConvNet-based 2D pose predictor to the subject under observation \cite{charles2016}. In contrast to its 2D counterpart, 3D human pose estimation suffers from the difficulty of gathering 3D groundtruth. While gathering large-scale 2D pose annotations from images is feasible, collecting corresponding 3D groundtruth is not. Instead, most works have relied on limited 3D annotations captured with  motion capture (MoCap) rigs in very restrictive indoor settings. Ideally, a simple, marker-less, multi-camera approach could provide reliable 3D human pose estimates in general settings. Leveraging these estimates as 3D annotations of images would capture the variability in users, clothing, and settings, which is crucial for ConvNets to properly generalize.

Towards this goal, this paper proposes a geometry-driven approach to automatically harvest reliable annotations from multi-view imagery. Figure~\ref{fig:overview1} provides an overview of our approach to automatically harvest reliable joint annotations. Given a set of images captured with a calibrated multi-view setup, a generic ConvNet for 2D human pose~\cite{newell2016stacked} produces single-view confidence heatmaps for each joint. The heatmaps in each view are backprojected to a common discretized 3D space, functioning as unary potentials of a 3D pictorial structure \cite{fischler1973,felzenszwalb2005}, while a tree graph models the pairwise relations between joints. The marginalized posterior distribution of the 3D pictorial structures model for each joint is used to identify which estimates are reliable. These reliable keypoints are used as annotations.

\begin{figure}[t]
	\begin{minipage}{1.00\textwidth}
	
	\begin{subfigure}{.5\textwidth}
	  \centering
	  \includegraphics[width=0.99\linewidth,trim={7cm 10cm 7cm 10cm},clip]{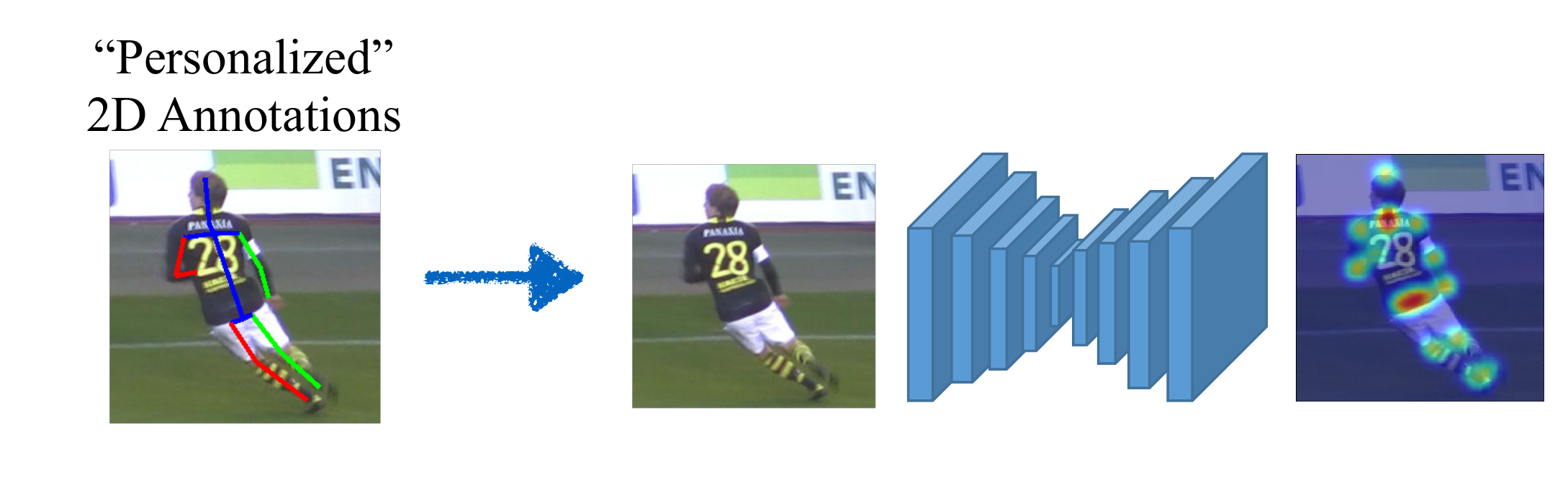}
	  \vspace{-1em}
	  \caption{``Personalizing'' a 2D pose ConvNet}
	  \label{fig:lever1}
	\end{subfigure}

	\begin{subfigure}{0.5\textwidth}
	  \centering
	  \includegraphics[width=0.99\linewidth,trim={7cm 10cm 7cm 10cm},clip]{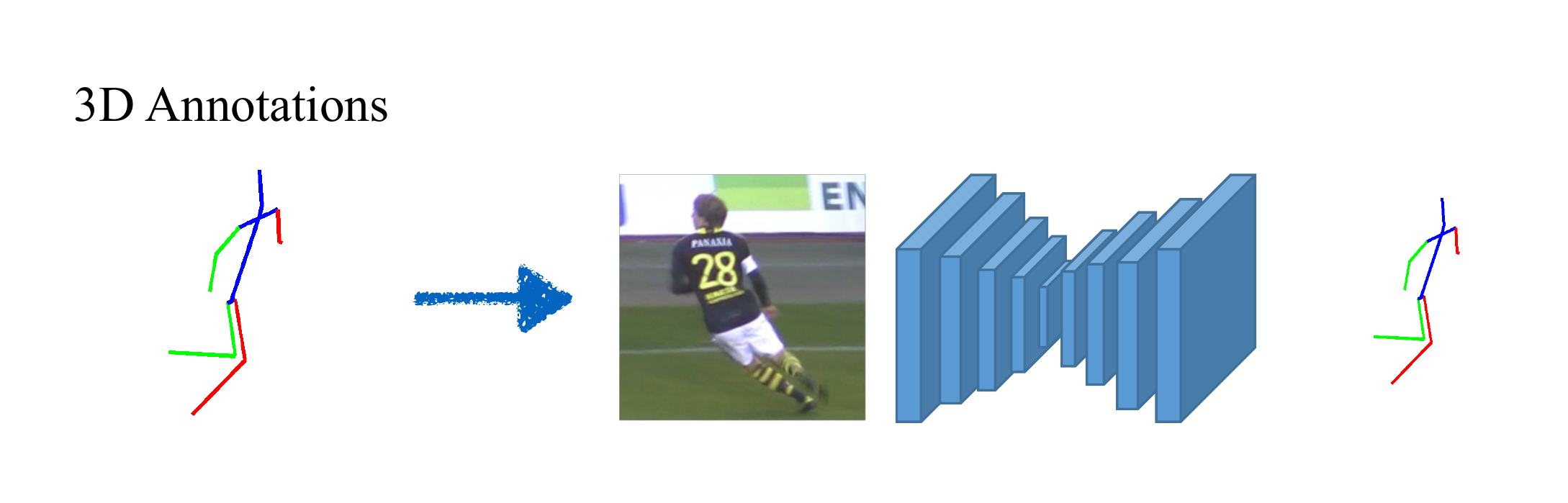}
	  \vspace{-1em}
	  \caption{Training a 3D pose ConvNet}
	  \label{fig:lever2}
	\end{subfigure}
	\end{minipage}
    \caption{
The quality of the harvested annotations is demonstrated in two applications: (a) projecting the 3D estimates into the 2D imagery and using them to adapt (``personalize'') a generic 2D pose ConvNet to the discriminative appearance aspects of the subject, (b) training a ConvNet that predicts 3D human pose from a single color image.
}
 \label{fig:overview2}
\end{figure}

Besides achieving state-of-the-art performance as compared to previous multi-view human pose estimators, our approach provides abundant annotations for pose-related learning tasks. In this paper, we consider two tasks. In the first task, we project the 3D pose annotations to the 2D images to create ``personalized'' 2D groundtruth, which is used to adapt the generic 2D ConvNet to the particular test conditions (Figure~\ref{fig:lever1}). In the second task, we use the 3D pose annotations to train from scratch a ConvNet for single view 3D human pose estimation that is on par with the current state-of-the-art.  Notably, in training our pose predictor, we limit the training set to the harvested annotations and do not use the available 3D groundtruth (Figure~\ref{fig:lever2}).

In summary, our four main contributions are as follows:
\begin{itemize}
\item We propose a geometry-driven approach to automatically acquire 3D annotations 
for human pose without 3D markers;
\item the harvested annotations are used to fine-tune a pre-trained ConvNet for 2D pose prediction to adapt to the discriminative aspects of the appearance of the subject under study, i.e., ``personalization''; we empirically show significant performance benefits;
\item  the harvested annotations are used to train from scratch a ConvNet that maps an image to a 3D pose, which is on par with the state-of-the-art, even though none of the available 3D groundtruth is used;
\item our approach for multi-view 3D human pose estimation achieves state-of-the-art results on standard benchmarks, which further underlines the effectiveness of our approach in exploiting the available multi-view information.
\end{itemize}

\section{Related work}

\noindent
{\bf Data scarcity for human pose tasks}: Chen \etal~\cite{chen2016synthesizing} and Ghezelghieh \etal~\cite{ghezelghieh2016learning} create additional synthetic examples for 3D human pose to improve ConvNet training. Rogez and Schmid~\cite{rogez2016mocap} introduce a collage approach. They combine human parts from different images to generate examples with known 3D pose. Yasin \etal~\cite{yasin2016dual} address the data scarcity problem, by leveraging data from different sources, e.g., 2D pose annotations and MoCap data. Wu \etal~\cite{wu2016single} also integrate dual source learning within a single ConvNet. Instead of creating synthetic examples, or bypassing the missing data, the focus of our approach is different. In particular, our goal is to gather images with corresponding 2D and 3D automatically generated annotations and use them to train a ConvNet. This way we employ images with statistics similar to those found in-the-wild,
which have been proven to be of great value for ConvNet-based approaches.

\vspace{3pt}
\noindent
{\bf 2D human pose}: Until recently, the dominant paradigm for 2D human pose involved local appearance modeling of the body parts coupled with the enforcement of structural constraints with a pictorial structures model~\cite{andriluka2009pictorial,yang2013,pishchulin2013poselet}. Lately though, end-to-end approaches using ConvNets have become the standard in this domain. The initial work of Toshev and Szegedy~\cite{toshev2014deep} regressed directly the $x,y$ coordinates of the joints using a cascade of ConvNets. Tompson \etal~\cite{tompson2014joint} proposed the regression of heatmaps to improve training. Pfister \etal~\cite{pfister2015flowing} proposed the use of intermediate supervision, with Wei \etal~\cite{wei2016convolutional} and Carreira \etal~\cite{carreira2015human} refining iteratively the network output. More recently, Newell \etal~\cite{newell2016stacked} built upon previous work to identify the best practices for human pose prediction and propose an hourglass module consisting of ResNet components~\cite{he2015deep}, and iterative processing to achieve state-of-the-art performance on standard benchmarks~\cite{andriluka20142d,sapp2013modec}. In this work, we employ the hourglass architecture as our starting point for generating automatic 3D human pose annotations.

\vspace{3pt}
\noindent
{\bf Single view 3D human pose}: 3D human pose estimation from a single image has been typically approached by applying more and more powerful discriminative methods on the image and combining them with expressive 3D priors to recover the final pose~\cite{simo2012single,zhou2016sparseness,bogo2016keep}. As in the 2D pose case, ConvNets trained end-to-end have grown in prominence. Li and Chan~\cite{li20143d} regress directly the $x,y,z$ spatial coordinates for each joint. Tekin \etal~\cite{tekin2016structured} additionally use an autoencoder to learn and enforce structural constraints on the output. Pavlakos \etal~\cite{pavlakos2016coarse} instead propose the regression of 3D heatmaps instead of 3D coordinates. Li \etal~\cite{li2015maximum} follow a nearest neighbor approach between color images and pose candidates. Rogez and Schmid~\cite{rogez2016mocap} use a classification approach, where the classes represent a sample of poses. To demonstrate the quality of our harvested 3D annotations, we also regress the $x,y,z$ joint coordinates~\cite{li20143d,tekin2016structured}, while employing a more recent architecture~\cite{newell2016stacked}.

\vspace{3pt}
\noindent
{\bf Multi-view 3D human pose}: Several approaches \cite{bergtholdt2010,amin2013,burenius2013,kazemi2013,belagiannis2014,belagiannis2015} have extended the pictorial structures model~\cite{fischler1973,felzenszwalb2005} to reason about 3D human pose taken from multiple (calibrated) viewpoints. Earlier work proposed simultaneously reasoning about 2D pose across multiple views, and triangulating 2D estimates to realize actual 3D pose estimates~\cite{bergtholdt2010,amin2013}. Recently, Elhayek~\etal~\cite{elhayek2015,elhayek2017marconi} used ConvNet pose detections for multi-view inference, but with a focus on tracking rather than annotation harvesting, as pursued here. Similar to the current paper, 3D pose has previously been directly modelled in 3D space \cite{burenius2013,kazemi2013,belagiannis2014,belagiannis2015}. A straightforward application of the basic pictorial structures model to 3D is computationally expensive due to the six degrees of freedom for the part parameterization. Our parameterization instead models only the 3D joint position, something that has also been proposed in the context of single view 3D pose estimation~\cite{kostrikov2014depth}. This instantiation of the pictorial structure makes inference tractable since we deal with three degrees of freedom rather than six.

\vspace{3pt}
\noindent
{\bf Personalization}: Consideration of pose in video presents an opportunity to tune the appearance model to the discriminative appearance aspects of the subject and thus improve performance. Previous work~\cite{ramanan2007}  leveraged this insight by using a generic pose detector to initially identify a set of high-precision canonical poses. These detections are then used to train a subject-specific detector. Recently, Charles \etal~\cite{charles2016} extended this idea using a generic 2D pose ConvNet to identify a select number of high precision annotations. These annotations are propagated across the video sequence based on 2D image evidence, e.g., optical flow. Regarding identifying confident predictions, the work of Jammalamadaka~\etal~\cite{jammalamadaka2012has} is related, where they extract features from the image and the output and train an evaluator to estimate whether the predicted pose is correct. In our work, rather than using 2D image cues to identify reliable annotations, our proposed approach leverages the rich 3D geometry presented by the multi-view setting and the constraints of 3D human pose structure, to combine and consolidate single view information. Such cues are highly reliable and complementary to image-based ones.

\section{Technical approach}\label{sec:technical}
The following subsections describe the main components of our proposed approach. Section~\ref{sec:generic} gives a brief description of the generic ConvNet used for 2D pose predictions. Section~\ref{sec:3dps} describes the 3D pictorial structures model used to aggregate multi-view image-driven keypoint evidence (i.e., heatmaps) provided as output by a ConvNet-based 2D pose predictor with 3D geometric information from a human skeleton model. Section~\ref{sec:selection} describes our annotation selection scheme that identifies reliable keypoint estimates based on the marginalized posterior distribution of the 3D pictorial structures model for each keypoint. The proposed uncertainty measure inherently integrates image evidence across all viewpoints and geometry. Finally, Sections \ref{sec:refinement} and \ref{sec:convnet3d} present two applications of our annotation harvesting approach. Section \ref{sec:refinement} describes the use of the harvested annotations to fine-tune an existing 2D pose ConvNet predictor. The resulting adapted predictor is sensitive to the discriminative aspects of the appearance of the subject under consideration, i.e., ``personalization''. Section \ref{sec:convnet3d} describes how we use the harvested annotations to train from scratch a 3D pose ConvNet predictor that maps a single image to 3D pose.
\subsection{Generic ConvNet} \label{sec:generic}

The initial component of our approach is a generic ConvNet for 2D human pose estimation that provides the initial set of noisy predictions for single view images. Since our approach is agnostic to the particular network architecture, any of the top-performing ConvNets is sufficient for this step, e.g.,~\cite{wei2016convolutional,bulat2016human,newell2016stacked}. Here, we adopt the state-of-the-art stacked hourglass design~\cite{newell2016stacked}. The main architectural component of this network is the hourglass module which consists of successive convolutional and pooling layers, followed by convolutional and upsampling layers, leading to a symmetric hourglass design. Stacking multiple hourglasses together allows for iterative processing of the image features. Best performance is achieved by the use of intermediate supervision, forcing the network to produce one set of predictions at the end of each hourglass. 

The prediction of the network is in the form of 2D heatmaps for each joint. The entire heatmap output includes useful information regarding the confidence of predictions, and can be considered as a 2D distribution of the joint locations. To take advantage of the entire heatmap prediction, we backproject the 2D distributions of the joints in a discretized 3D cube. This is used to accommodate the predictions for all the views and serves as the inference space for 3D pictorial structures model, described in Sec.~\ref{sec:3dps}.

\subsection{Multi-view optimization} \label{sec:3dps}

The pose model used to aggregate information across views is based on a 3D generalization of the classical pictorial structures model~\cite{fischler1973,felzenszwalb2005}. A major departure of the current work from prior 3D instantiations of multi-view approaches (e.g., \cite{burenius2013}) is the use of a joint representation, $\mathbf{S} = \{\mathbf{s}_i | i=1,\dotsc,N \}$, where $\mathbf{s}_i\in \mathbb{R}^3$ encodes the 3D position of each joint, rather than the 3D configuration of parts, i.e., limbs. The simplified parameterization and tree structure for the pairwise terms admit efficient 3D joint configuration inference via dynamic programming, i.e., the sum-product algorithm.

\vspace{3pt}
\noindent
{\bf Articulation constraints}: The pairwise relation between joints is modelled by a tree structure of the human skeleton.  The edge set is denoted by $\mathcal{E}$ and the edge (i.e., limb) lengths by $\{L_{ij}|(i,j)\in \mathcal{E}\}$. The prior distribution is given by the articulation constraints and can be written as
\begin{align}
p(\mathbf{S}) \propto \prod_{(i,j)\in \mathcal{E}} p(\mathbf{s}_i,\mathbf{s}_j).
\end{align}
The pairwise terms, $p(\mathbf{s}_i,\mathbf{s}_j)$, constrain the lengths of the human limbs $L_{ij}$:
\begin{align}
p(\mathbf{s}_i,\mathbf{s}_j) =
\begin{cases}
    1, & \text{if } L_{ij}-\varepsilon \leq \|\mathbf{s}_i-\mathbf{s}_j\| \leq L_{ij}+\varepsilon\\
    0, & \text{otherwise}
\end{cases},
\end{align}
where $\varepsilon = 1$ is used as a tolerance for the variability from the expected limb length $L_{ij}$ of the subject. More sophisticated pairwise terms can also be adopted if MoCap data are available, e.g.,~\cite{kostrikov2014depth}.

\vspace{3pt}
\noindent
{\bf Data likelihood}: Given a 3D pose, the likelihood of seeing $M$ synchronized images from $M$ calibrated cameras is modeled as
\begin{align}
p(\mathbf{I}|\mathbf{S}) \propto \prod_{k=1}^{M}\prod_{i=1}^{N} p(\mathbf{I}_k|\pi_k(\mathbf{s}_i)),
\end{align}
where $\pi_k(\mathbf{s}_i)$ denotes the 2D projection of $\mathbf{s}_i$ in the $k$-th view given the camera parameters. The data likelihood, $p(\mathbf{I}_k|\pi_k(\mathbf{s}_i))$, is modelled by the multi-channel heatmap outputs of the ConvNet (Sec.\ \ref{sec:generic}).

\vspace{3pt}
\noindent
{\bf Inference}: Finally, the posterior distribution of a 3D pose given 2D images from different views is given by:
\begin{align}
p(\mathbf{S}|\mathbf{I}) \propto \prod_{k=1}^{M}\prod_{i=1}^{N} p(\mathbf{I}_k|\pi_k(\mathbf{s}_i)) \prod_{(i,j)\in \mathcal{E}} p(\mathbf{s}_i,\mathbf{s}_j).
\end{align}
The solution space of the 3D joint position is restricted to a 3D bounding volume around the subject and quantized by a $64\times64\times64$ grid. Pose estimates are computed as the mean of the marginal distribution of each joint given the multi-view images. The marginal distribution of the discrete variables is efficiently computed by the sum-product algorithm~\cite{felzenszwalb2005}.

\subsection{Annotation selection} \label{sec:selection}

The 3D reconstructions provided by the multi-view optimization offer a very rich but noisy set of annotations. We are effectively equipped with automatic 3D annotations for all the images of the multi-view setup. Moreover, these annotations integrate appearance cues from each view (2D pose heatmaps), geometric constraints from the multiple views (backprojection in a common 3D space), as well as constraints from the articulated structure (3D pictorial structure). This allows us to capitalize on the available information from the images and the 3D geometry to provide a robust set of annotations.

For further benefits, we proceed to a selection step over the annotations provided from the 3D reconstruction. A useful property of our multi-view optimization using the pictorial structures model is that the marginalized distribution of each joint offers a measure of the prediction's uncertainty. This means that we are provided with a selection cue for free. For example, the determinant of the 3D covariance matrix for each joint's marginalized distribution can be used as a confidence measure to decide whether the joint will be used as an annotation. In our experiments, we identify as reliable annotations the $70\%$ most confident predictions for each joint in terms of the determinant of the 3D covariance matrix, although other measures are also possible.

\subsection{``Personalizing'' 2D pose ConvNet} \label{sec:refinement}

The goal of ``personalization'' is to adapt the original detector such that it captures the discriminative appearance aspects of the subject of interest, such as clothing. Both Ramanan~\etal~\cite{ramanan2007} and Charles \etal~\cite{charles2016} proposed methods to ``personalize'' a detector using 2D evidence (e.g., optical flow) from monocular video. Instead, our proposed approach focuses on cues provided by image evidence, geometric properties of the multi-view setup, and structural constraints of the human body.

Given the set of selected annotations, we use them to fine-tune a generic 2D pose ConvNet with backpropagation, such that it adapts to the testing conditions of interest. The procedure is very similar to the one used to train the ConvNet in the first place, with the difference that we leverage our automatically generated annotations as targets for the available images. The target heatmaps consist of a 2D Gaussian with a standard deviation $\sigma = 1$ pixel, centered on the annotation location of the joint. A separate heatmap is synthesized for each joint. During training, we use a Mean Squared-Error loss between the predicted and the target heatmaps. If the joint is not within the selected annotation set (i.e., the localization is not confident), we simply ignore the loss incurred by it during optimization. We terminate refinement after four epochs through our auto-annotated data to avoid overfitting on the given examples.

\subsection{3D pose ConvNet training} \label{sec:convnet3d}

For 3D human pose estimation, we train a ConvNet from scratch that takes a single image as input and predicts the 3D pose. Our formulation follows the coordinate regression paradigm~\cite{li20143d,tekin2016structured}, but more sophisticated methods can also be employed, e.g., the volumetric representation for 3D pose~\cite{pavlakos2016coarse}. The target of the network is the $x,y,z$ coordinates of $N$ human body joints. For $x,y$ we use pixel coordinates, while $z$ is expressed in metric depth with respect to a specified root joint (here the pelvis is defined as the root). We organize the output in a single $3N$-dimensional vector. The network is supervised with an $\mathcal{L}_2$ regression loss:
\begin{eqnarray}
\mathcal{L} = \sum_{n=1}^N \| \bm{x}_{gt}^n - \bm{x}_{pr}^n \|_2^2,
\end{eqnarray}
where $\bm{x}_{gt}^n$ is the groundtruth and $\bm{x}_{pr}^n$ is the predicted location for joint $n$. The architecture we use is a single hourglass module~\cite{newell2016stacked} with the addition of a fully connected layer at the end to allow every output to have a connection with each activation of the previous feature volume.

\section{Empirical evaluation} \label{sec:experiments}

This section is dedicated to the empirical evaluation of our proposed approach. First, we give a description of the datasets used (Section~\ref{sec:datasets}). Next, we briefly discuss the implementation details of our approach (Section~\ref{sec:exp-details}). Finally, we present the quantitative (Sections~\ref{sec:exp-multi} to~\ref{sec:exp-single}) and the qualitative evaluations (Section~\ref{sec:qualitative}).

\subsection{Datasets} \label{sec:datasets}

For our quantitative evaluation we focused on two datasets that target human pose estimation and provide a multiple camera setup; (i) KTH Multiview Football II~\cite{burenius2013}, a small-scale outdoor dataset with challenging visual conditions, and (ii) Human3.6M~\cite{ionescu2014human}, a large-scale indoor dataset, with a variety of available scenarios.

\textbf{KTH Multiview Football II} \cite{burenius2013} contains images of professional footballers playing a match. Evaluation for 3D pose was performed using the standard protocol introduced with the dataset \cite{burenius2013} and used elsewhere \cite{kazemi2013,belagiannis2015}, where Sequence 1 of ``Player 2'' is used for testing.  Reported results are based on the percentage of correct parts (PCP) to measure 3D part localization using the two and three camera setups. Additional evaluation for 2D pose was performed using Sequence 2 of ``Player 2'' for testing, where reported results are based on the percentage of correct parts in 2D.

\textbf{Human3.6M}  \cite{ionescu2014human} is a recent large-scale dataset for 3D human sensing captured in a lab setting. It includes 11 subjects performing 15 actions, such as walking, sitting, and phoning. Following previous work \cite{li2015maximum,zhou2016sparseness}, we use two subjects for testing (S9 and S11), and report results based on the average 3D joint error.

It is crucial to mention that in the experiments presented below, no groundtruth data was leveraged for training from the respective datasets. We relied solely on the generic 2D ConvNet (trained on MPII~\cite{andriluka20142d}) and the knowledge of the geometry from the calibrated camera setup.

\subsection{Implementation details} \label{sec:exp-details}

For the generic 2D pose ConvNet, we use a publicly available model~\cite{newell2016stacked}, which is trained on the MPII human pose dataset~\cite{andriluka20142d}. To ``personalize'' a given 2D pose ConvNet through fine-tuning, we maintain the same training details as the ones described in the original work. The learning rate is set to 2.5e-4, the batch size is 4, rmsprop is used for optimization and data augmentation is used, that includes rotation ($\pm 30^o$), scale ($\pm 0.25$),
and left-right flipping.

To train the 3D pose ConvNet, we employ the same architecture, but we use only one hourglass component and add a fully connected layer at the end, to regress the $N$ joints coordinates. The training details with respect to optimization and data augmentation are the same as for the initial network, but training is done from scratch (we do not use a pretrained model).

\subsection{Multi-view pose estimation}  \label{sec:exp-multi}

First of all, we need to assess the accuracy of the annotations provided from our multi-view optimization scheme. Since our ConvNets are not trained using groundtruth data from the aforementioned datasets, we heavily rely on the quality of these automatic annotations. Therefore, we evaluate multi-view pose estimation using our approach, described in Section~\ref{sec:3dps}

First, we report results of our approach on the small-scale, yet challenging KTH dataset. Even though relevant methods train specialized 2D detectors for pose estimation, they are all outperformed by our approach using only a generic ConvNet for 2D joint prediction. The relative improvement is illustrated in Table~\ref{table:kth_multi}.

\begin{table}[!t]
\small
\begin{center}
\setlength{\tabcolsep}{1.5pt}
\begin{tabular}{|r|c|c|c|c||c|c|c|c|c|}\cline{2-10}
\multicolumn{1}{c}{}&\multicolumn{4}{|c||}{Two cameras} & \multicolumn{5}{c|}{Three cameras}\\\cline{2-10}
\multicolumn{1}{c|}{}& \multirow{1}{*}{\cite{burenius2013}} & \multirow{1}{*}{\cite{belagiannis2014}} &\multirow{1}{*}{\cite{belagiannis2015}}  & Ours   & \multirow{1}{*}{\cite{burenius2013}}& \multirow{1}{*}{\cite{kazemi2013}}& \multirow{1}{*}{\cite{belagiannis2014}} &\multirow{1}{*}{\cite{belagiannis2015}}&Ours\\\hline
Upper arms 	& 53 	& 64		& 96		& {\bf 98} 	& 60			& 89			& 68		& 98		&{\bf 100} \\
Lower arms	& 28 	& 50		& 68		& {\bf 92}	& 35			& 68			& 56		& 72		&{\bf 100} \\
Upper legs	& 88		& 75		& 98		& {\bf 99}	& {\bf 100}	& {\bf 100}	& 78		& 99		&{\bf 100} \\
Lower legs	& 82 	& 66		& 88		& {\bf 97}	& 90			& 99			& 70		& 92		&{\bf 100} \\\hline\hline
Average 		& 62.7 	& 63.8	& 87.5	& {\bf 96.5}	& 71.2		& 89.0		& 68.0	& 90.3	&{\bf 100} \\\hline
\end{tabular}
\end{center}
\vspace{-15pt}
\caption{Quantitative comparison of multi-view pose estimation methods on KTH Multiview Football II. The numbers are the percentage of correct parts (PCP) in 3D using two and three cameras. Baseline numbers are taken from the respective papers. In constrast to the compared methods, no training data from this dataset was used for our approach.}
\label{table:kth_multi}
\end{table}

\begin{table*}[t]
\begin{center}
\setlength{\tabcolsep}{2pt}
\begin{tabular}{rcccccccc}\cline{1-9}
 & Directions & Discussion & Eating & Greeting & Phoning & Photo & Posing & Purchases \\\hline
Li et al. \cite{li2015maximum} & - & 134.13 & 97.37 & 122.33 & - & 166.15 & - & - \\ 
Zhou et al. \cite{zhou2016sparseness} & 87.36 & 109.31 & 87.05 & 103.16 & 116.18 &  143.32& 106.88 & 99.78 \\
Tekin et al. \cite{tekin2016structured}  &        - &        129.06 &        91.43 &        121.68 &        - &        162.17 &        - &        - \\
Zhou et al. \cite{zhou2016deep}  &        91.83 &        102.41 &        96.95 &        98.75 &        113.35 &        125.22 &       90.04 &        93.84 \\
Ours  & {\bf 41.18} &        {\bf 49.19} &        {\bf 42.79} &        {\bf 43.44} &        {\bf 55.62} &        {\bf 46.91} &        {\bf 40.33} &        {\bf 63.68} \\\hline
 &  Sitting & SittingDown &  Smoking & Waiting & WalkDog &  Walking & WalkTogether &  Average \\\hline
Li et al. \cite{li2015maximum} & - & - & - & - & 134.13 & 68.51 & - & - \\
Zhou et al. \cite{zhou2016sparseness} & 124.52 & 199.23 & 107.42 & 118.09 & 114.23 & 79.39 & 97.70 & 113.01 \\
Tekin et al. \cite{tekin2016structured}  &        - &       - &        - &        - &        130.53 &       65.75 &        - &        - \\
Zhou et al. \cite{zhou2016deep}  &        132.16 &        158.97 &        106.91 &        94.41 &        126.04 &        79.02 &        98.96 &        107.26 \\
Ours  & {\bf 97.56} &        {\bf 119.90} &        {\bf 52.12} &        {\bf 42.68} &        {\bf 51.93} &        {\bf 41.79} &        {\bf 39.37} &        {\bf 56.89} \\\hline
\end{tabular}
\end{center}
\vspace{-15pt}
\caption{Quantitative evaluation of our approach on Human3.6M. The numbers are the average 3D joint errors (mm). Baseline numbers are taken from the respective papers. Note that Zhou~\etal~\cite{zhou2016sparseness} use video, while our proposed method is multi-view.
}
\label{table:hm36m_multi}
\end{table*}

For Human3.6M we apply the same method to multi-view pose estimation. Since this dataset was published very recently, there are no reported results for multi-view pose estimation methods. It is interesting though to compare with the top-performing works for single view 3D pose such that we can quantify the current gap between single view and multi-view estimation. The full results are presented in Table~\ref{table:hm36m_multi}. Our approach reduces the error of the state-of-the-art single view approach of Zhou \etal~\cite{zhou2016deep} by almost a half. We note that Zhou \etal~\cite{zhou2016sparseness} use video instead of prediction from a single frame. We do not include results from Bogo \etal~\cite{bogo2016keep} and Sanzari \etal~\cite{sanzari2016bayesian} which report average errors of $82.3$mm and $93.15$mm, respectively, since they use a rigid alignment between the estimated pose and the groundtruth, making it not comparable with the other methods. Moreover, as a weak multi-view baseline, we averaged the per view 3D estimates from one of the state-of-the-art approaches~\cite{zhou2016sparseness}. This naive combination achieves an average error of $103.10$mm which is a minimal improvement compared to the original error of $113.01$mm for the corresponding single view approach. This demonstrates that handling the views independently and combining the single view 3D pose results in a late stage does not leverage the rich 3D geometric constraints available and significantly underperforms compared to  our multi-view optimization. 

\subsection{``Personalizing'' 2D pose ConvNet}  \label{sec:exp-refine}

Having validated the accuracy of our proposed multi-view optimization scheme, the next step is to actually leverage the automatic annotations for learning purposes. The most immediate benefit comes from using them to refine the generic ConvNet and adapt it to the particular test conditions. This can be considered as an application of ``personalization'', similar in nature to the goal of Charles \etal~\cite{charles2016}, where significant pose estimation gains in terms of accuracy were reported.

For KTH we use the two available sequences from ``Player 2'' to evaluate the online adaptation of our network. Since our focus is to purely evaluate the quality of the 2D predictions before and after refinement, we report 2D PCP results in Table~\ref{table:kth_refine}. We observe performance improvement across all parts of the subject. Moreover, for the second sequence which is considerably more challenging, the benefit from our refinement is even greater. This underlines the importance of refinement when the original detector fails.

\begin{table}
\begin{center}
\setlength{\tabcolsep}{2pt}
\begin{tabular}{|r|c|c||c|c|}\cline{2-5}
\multicolumn{1}{c}{}&\multicolumn{2}{|c||}{Seq 1} & \multicolumn{2}{c|}{Seq2}\\\cline{2-5}
\multicolumn{1}{c|}{}& Generic & Refined & Generic & Refined\\\hline
Upper arms 		& 98 	& {\bf 100} 	& 80 	& {\bf 89} \\ 
Lower arms		& 89 	& {\bf 92} 	& 64		& {\bf 74} \\ 
Upper legs		& 98		& {\bf 100} 	& 85		& {\bf 91} \\ 
Lower legs		& 96 	& {\bf 100} 	& 79		& {\bf 86} \\\hline 
Average 			& 95.3 	& {\bf 98.0}	& 77.0	& {\bf 84.5} \\\hline 
\end{tabular}
\end{center}
\vspace{-15pt}
\caption{Quantitative comparison of the generic ConvNet versus the refined version for the two sequences of ``Player 2'' from KTH Multiview Football II. The numbers are percentage of correct parts (PCP) in 2D. Performance improvement is observed across all parts.}
\label{table:kth_refine}
\end{table}

For Human3.6M we evaluate the quality of 2D heatmaps through their impact on the multi-view optimization. Achieving better results for 2D pose estimation is definitely desirable, but ideally, the predicted heatmaps should benefit other post-processing steps as well, e.g., our multi-view optimization. In Table~\ref{table:hm36m_refine}, we provide a more detailed ablative study comparing different sets of annotations for refinement. Starting with the ``Generic'' ConvNet, one naive approach we compare against is using the heatmap maximum predictions as annotations (``HM''), or a subset of the most confident of those predictions (``HM+sel''). For ``HM+sel'' we use the heatmap value to indicate detection confidence, and identify only the top $70\%$ for each joint as reliable 2D annotations. These serve as baselines for refining the ConvNet. We also employ the complete annotation set that is provided from our multi-view optimization (``PS''), and a high quality version of this by selecting the most confident joint predictions only (denoted as ``PS+sel'' and described in Section~\ref{sec:selection}). The reported results include both the average performance across all 15 actions, as well as the performance on the three actions with the highest error, according to Table~\ref{table:hm36m_multi}, namely, Purchases, Sitting, and Sitting Down. Again, the performance benefits are greater for more challenging actions, which justifies the use of our method to overcome dataset bias and adapt to the scenario of interest. Also, the naive approach to recover more 2D annotations and bootstrapping on the output of the generic ConvNet (``HM'' and ``HM+sel'') is only marginally helpful on average, which underlines the benefit of the rich geometrical information we employ to recover annotations. Finally, the proposed selection scheme (``PS+sel'') outperforms the model that uses all annotations of the multi-view optimization (``PS'') which exemplifies the importance of selecting only a high-quality subset of the annotations for refinement.

\begin{table}[t]
\begin{center}
\setlength{\tabcolsep}{2pt}
\begin{tabular}{rcccc}\cline{1-5}
 \multicolumn{1}{c}{} & \multirow{2}{*}{Purchases} &  \multirow{2}{*}{Sitting} & Sitting 	&  Average \\
 \multicolumn{1}{c}{} &  											& 											& Down 	&  (15 actions)  \\\hline
Generic		& 63.68			& 97.56 				&	119.90 			& 56.89 	\\
HM			&	57.57			& 86.37 				& 	100.39				& 55.13 	\\
HM+sel		& 52.50			& 91.49 				& 110.30				& 	55.62	\\
PS				& 51.32			& 79.39				&	97.26				&	51.18	\\
PS+sel		& {\bf 45.98}	& {\bf 68.09} 		& {\bf 73.91}		& 	{\bf 47.83}	\\
\hline
\end{tabular}
\end{center}
\vspace{-15pt}
\caption{Quantitative comparison of multi-view optimization after fine-tuning the ConvNet with different annotation sets and evaluating on Human3.6M. We present results for the three most challenging actions (based on Table~\ref{table:hm36m_multi}), along with the average across all actions. The numbers are the average 3D joint error (mm). ``Generic'', ``HM'',  ``HM+sel'', ``PS'' and ``PS+sel'' are defined in Section~\ref{sec:exp-refine}.
}
\label{table:hm36m_refine}
\end{table}

\subsection{Training a 3D pose ConvNet}  \label{sec:exp-single}

A great challenge, but also a very interesting application of our method is to use the automatically generated annotations to train a ConvNet for 3D pose estimation. Since KTH is a small-scale dataset, we focus on Human3.6M. We leverage the high-quality annotations from the multi-view optimization scheme, and train the network described in Section~\ref{sec:convnet3d} from scratch. The results are presented in Table~\ref{table:hm36m_single}, along with other approaches. Even though we only use the noisy annotations recovered by our approach for training and ignored the groundtruth from the dataset, the final trained ConvNet is on par with the state-of-the-art.

\begin{table}[t]
\begin{center}
\setlength{\tabcolsep}{2pt}
\begin{tabular}{rcc}\cline{1-3}
 \multicolumn{1}{c}{} & Average (6 actions) & Average (15 actions)  \\\hline
Li \etal \cite{li2015maximum} 				& 121.31 & - \\
Tekin \etal \cite{tekin2016structured} 		& 116.77 & - \\
Park \etal \cite{park20163d}  				& 111.12 & 117.34 \\  
Zhou \etal \cite{zhou2016deep}  				& 104.73   & 107.26 \\
Rogez \etal~\cite{rogez2016mocap}			& - & 121.2 \\
Ours  & 113.65 & 118.41 \\\hline
\end{tabular}
\end{center}
\vspace{-12pt}
\caption{Quantitative comparison of single image approaches on Human3.6M. The numbers are the average 3D joint errors (mm). Baseline numbers are taken from the respective papers. In contrast to the other works, we do not use 3D groundtruth for training, instead we rely solely on the harvested 3D annotations. Despite that, our performance is on par with the state-of-the-art.
}
\label{table:hm36m_single}
\end{table}

\subsection{Qualitative results}  \label{sec:qualitative}

\begin{figure*}[t]
\includegraphics[width=0.33\textwidth]{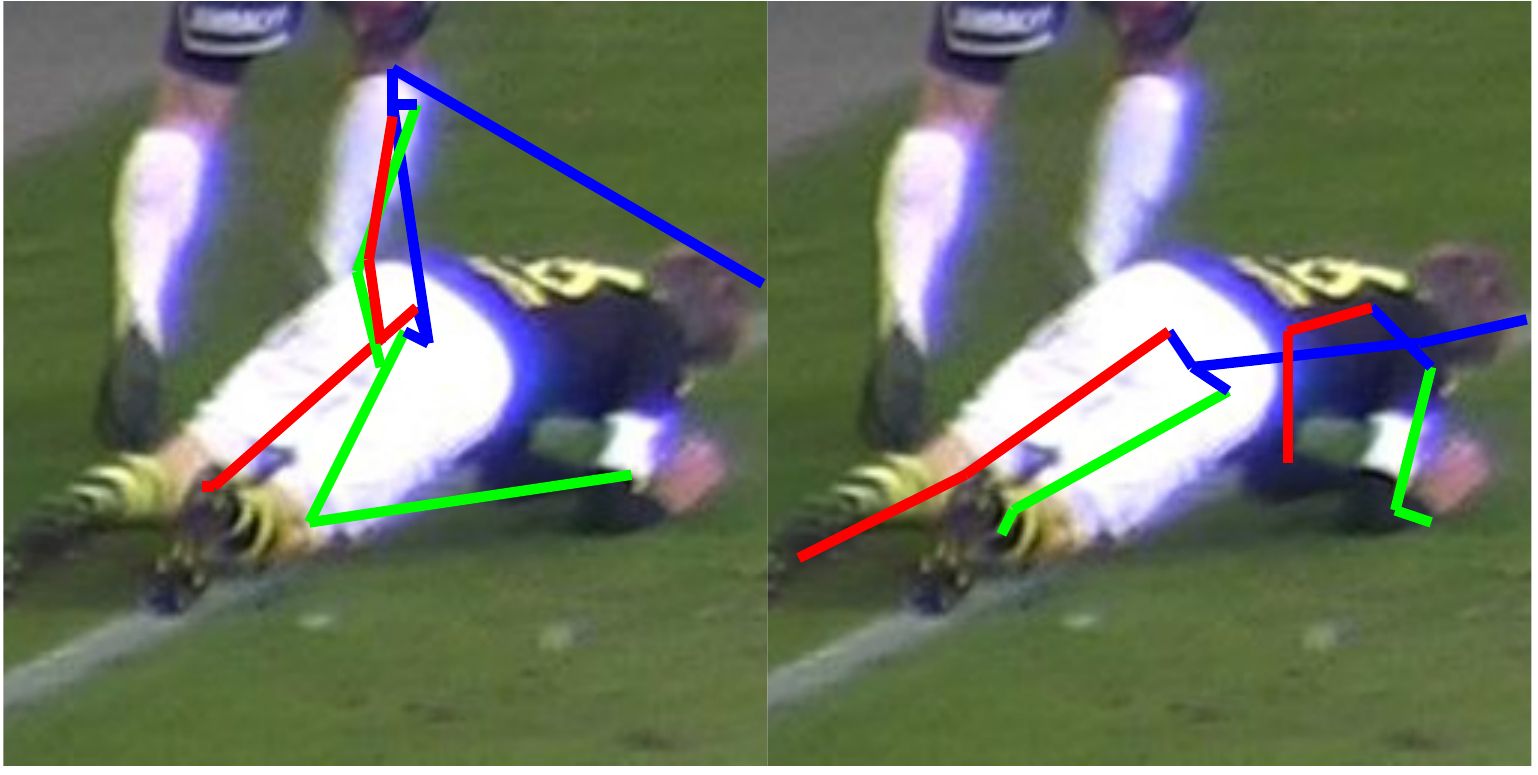} 
\includegraphics[width=0.33\textwidth]{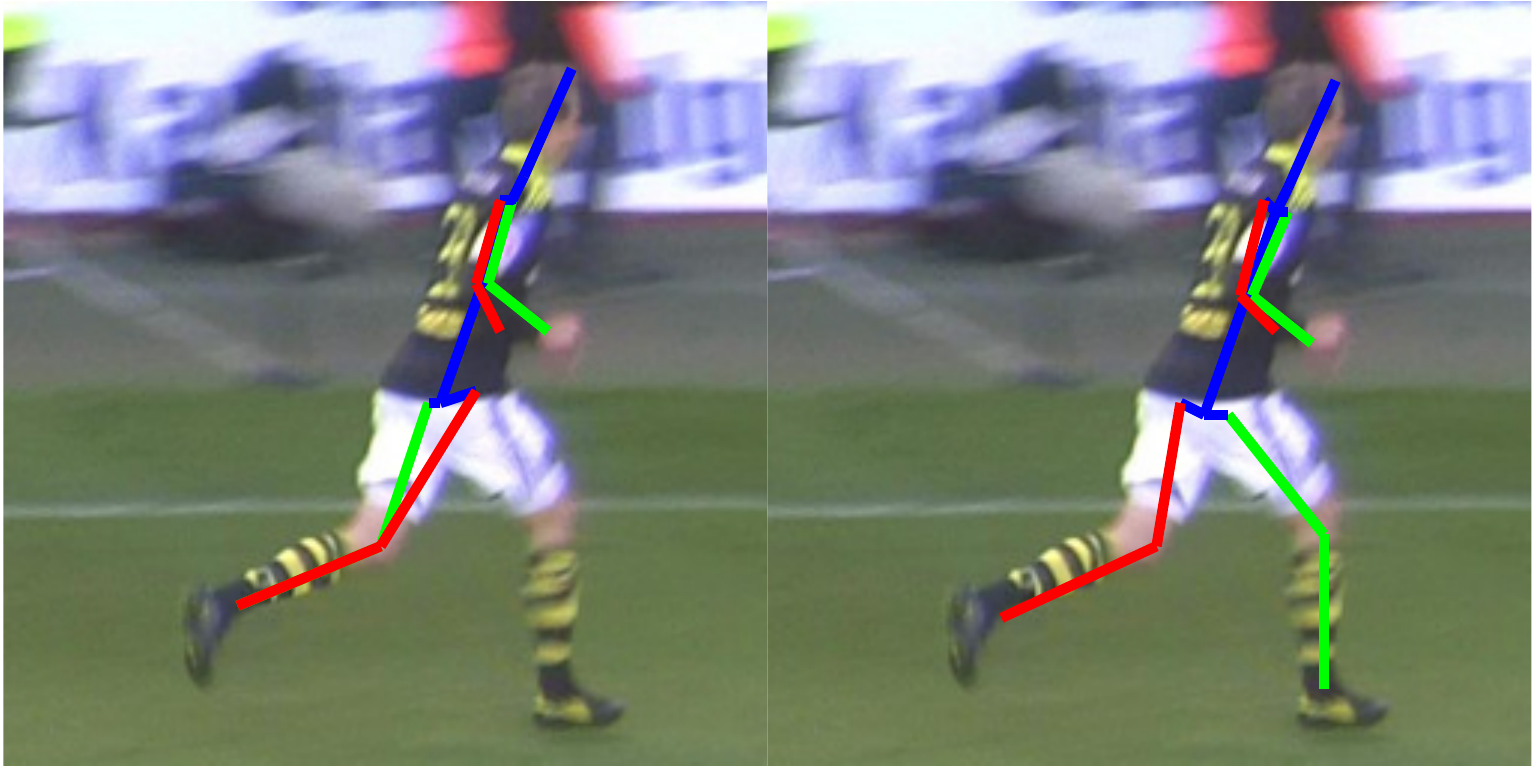} 
\includegraphics[width=0.33\textwidth]{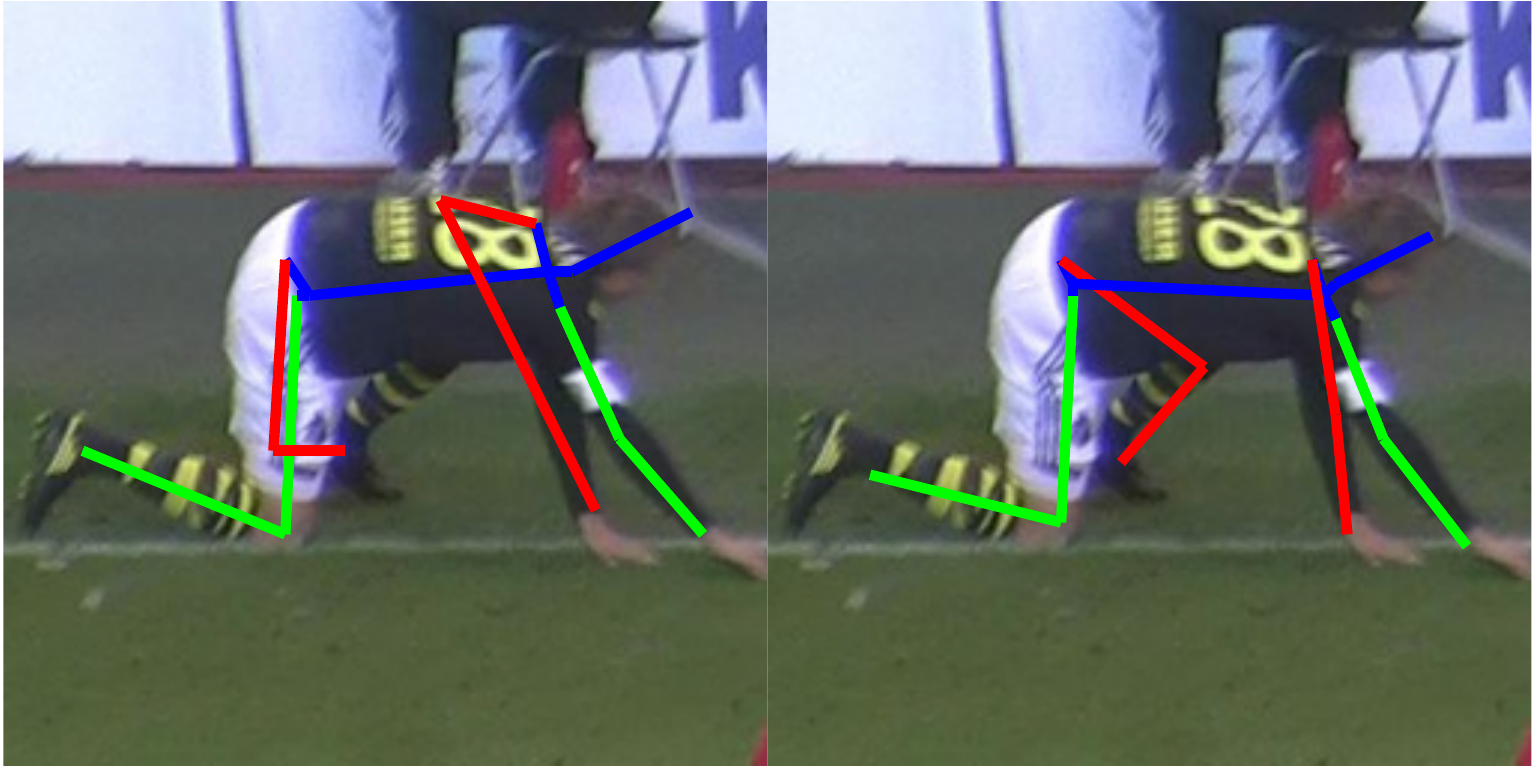}
\vspace{-8pt}
\caption{Examples on KTH Multiview Football II showing the performance gain from ``personalization''. For each pair of images, pose estimation results are presented from the generic (left) and the ``personalized'' ConvNet (right).}
\label{fig:KTH_qual}
\end{figure*}

\begin{figure*}[t]
\includegraphics[width=0.33\textwidth]{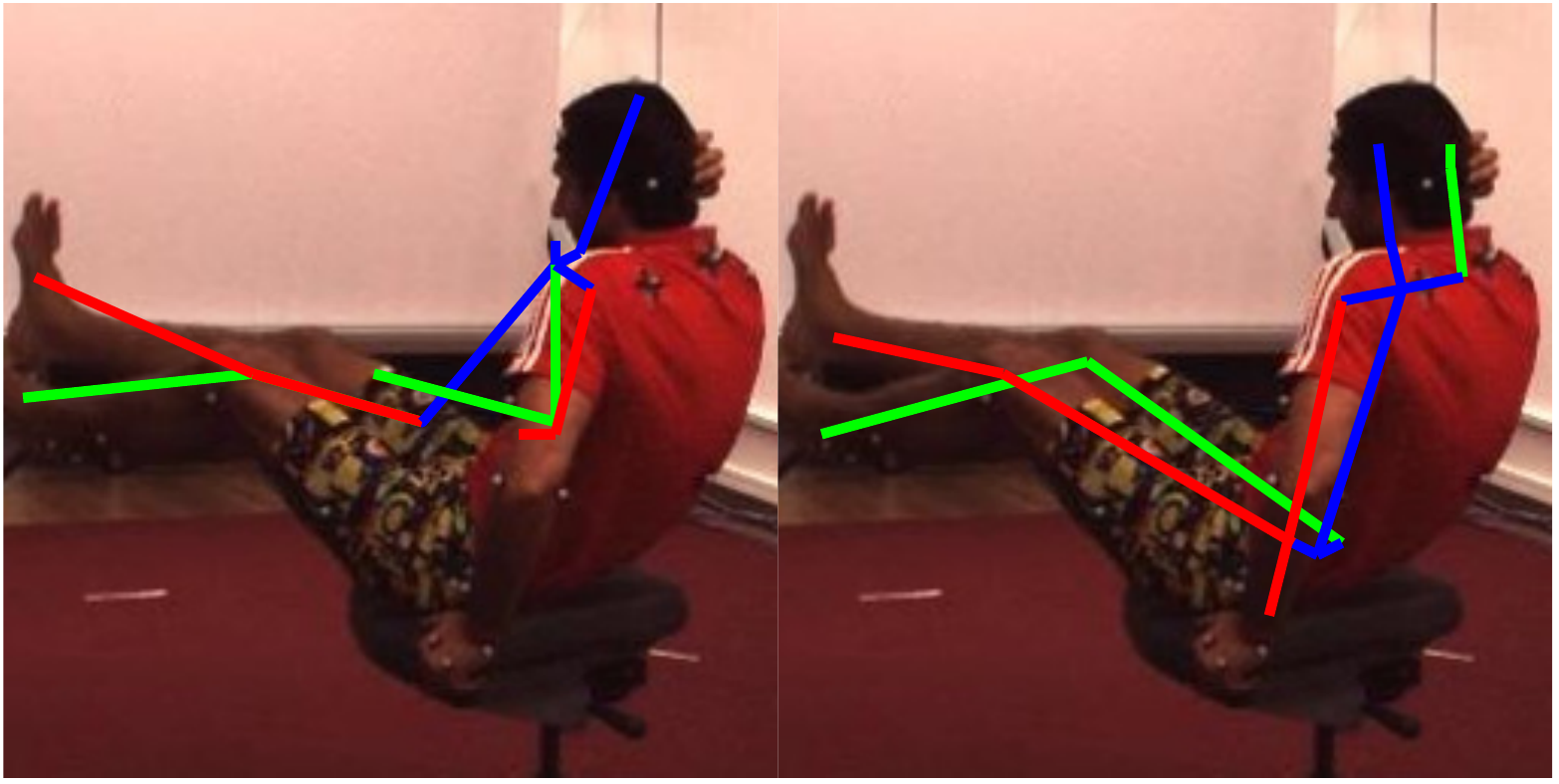}
\includegraphics[width=0.33\textwidth]{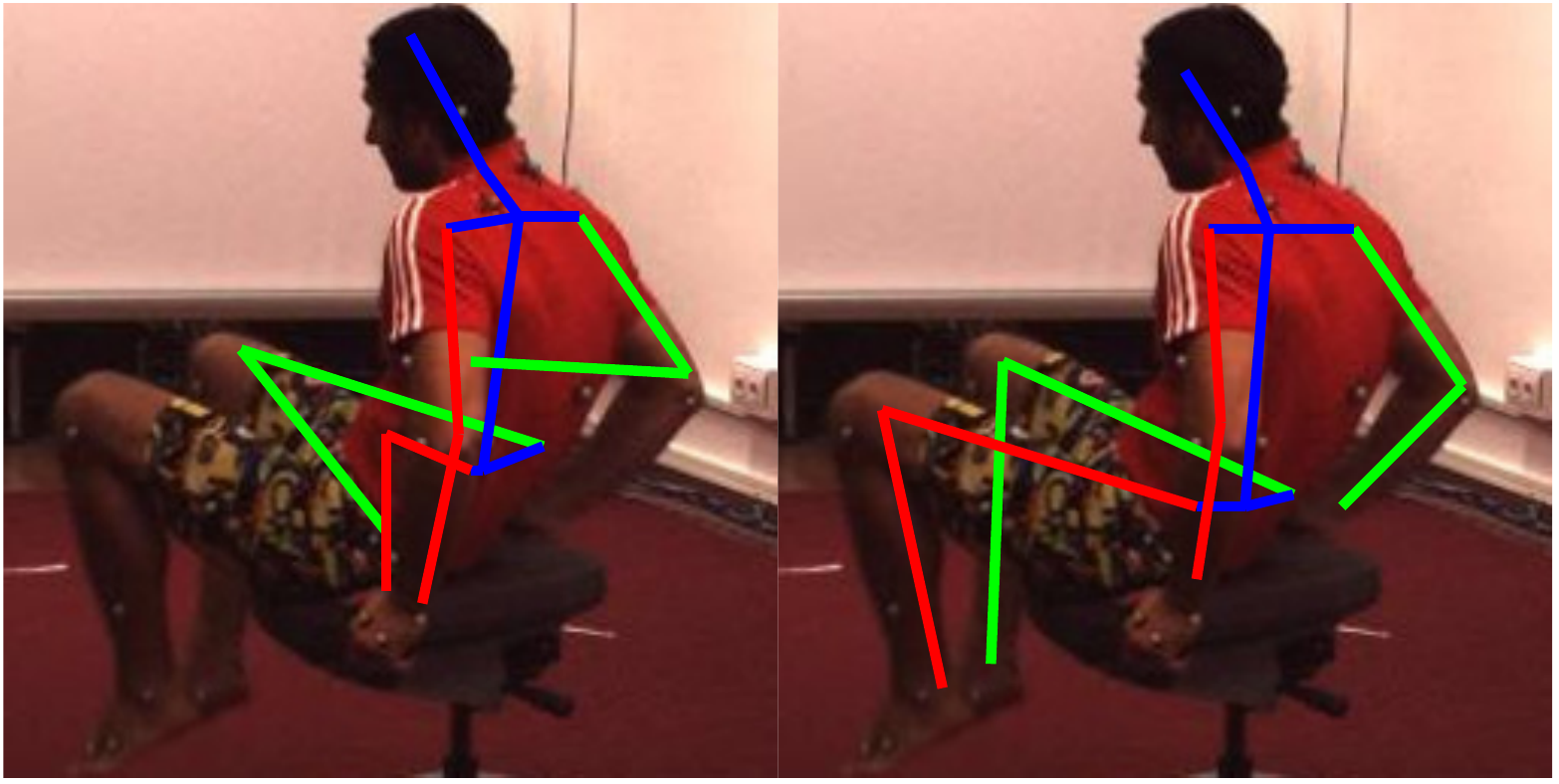}
\includegraphics[width=0.33\textwidth]{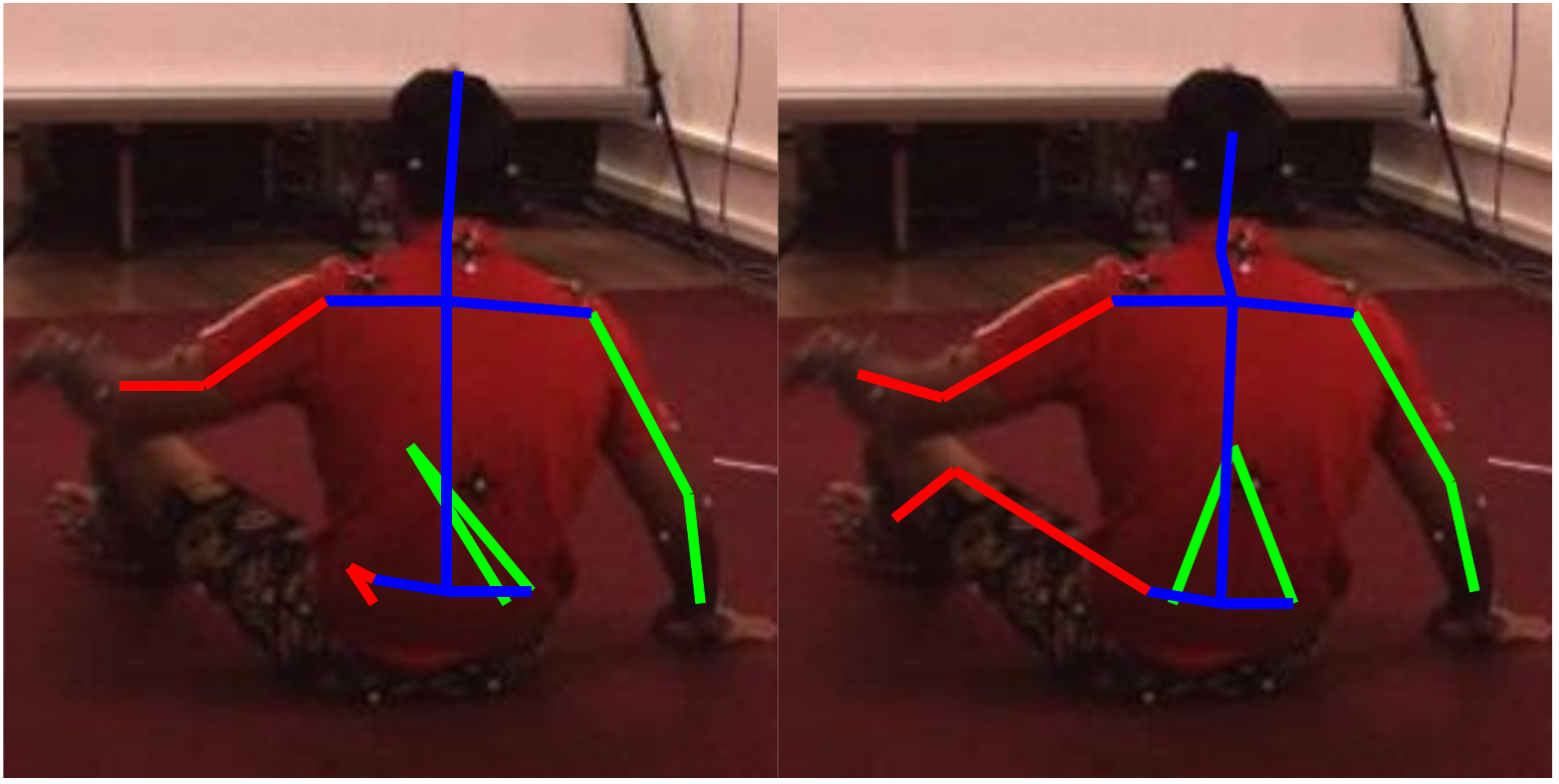} \\
\includegraphics[width=0.33\textwidth]{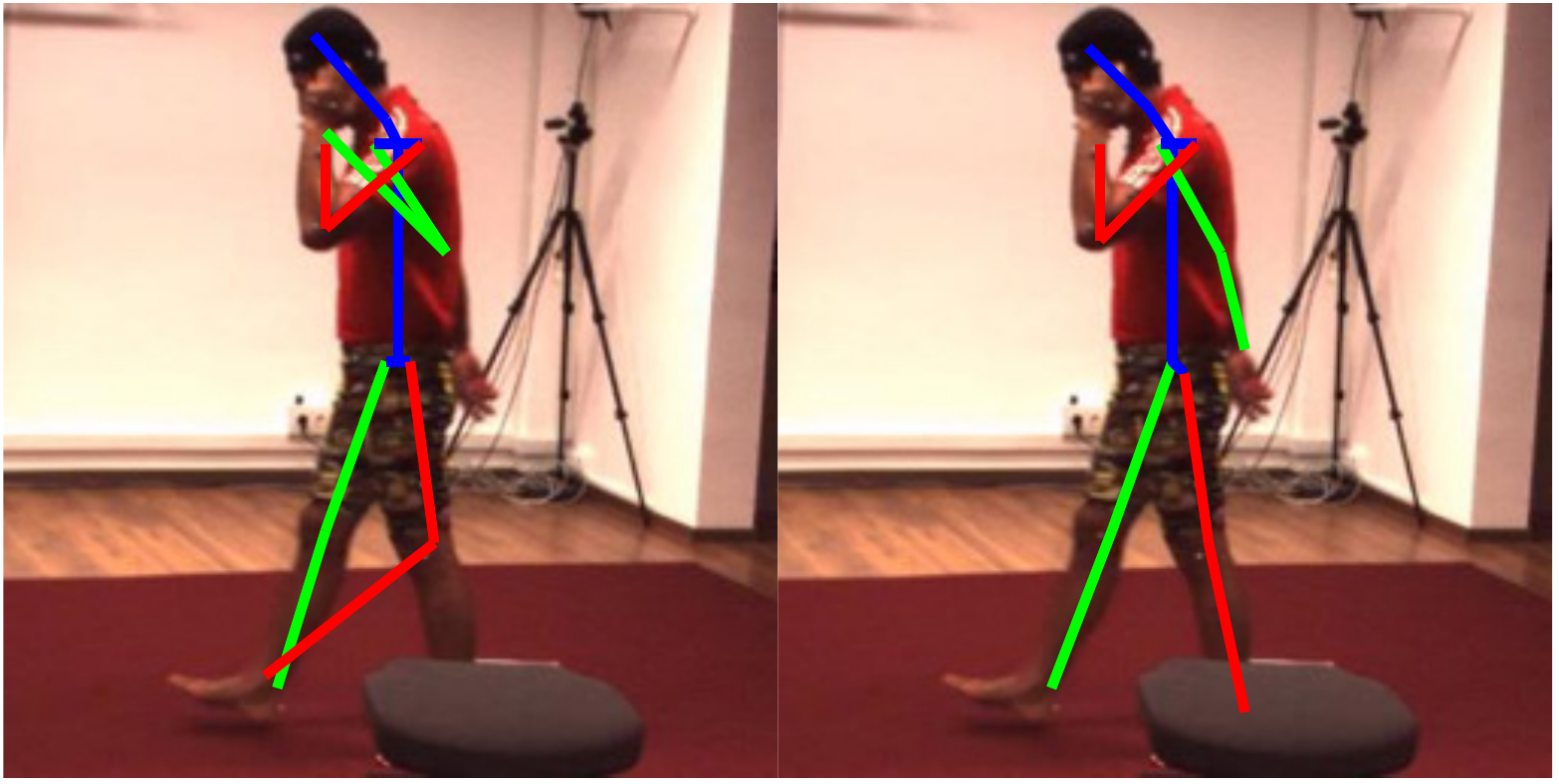}
\includegraphics[width=0.33\textwidth]{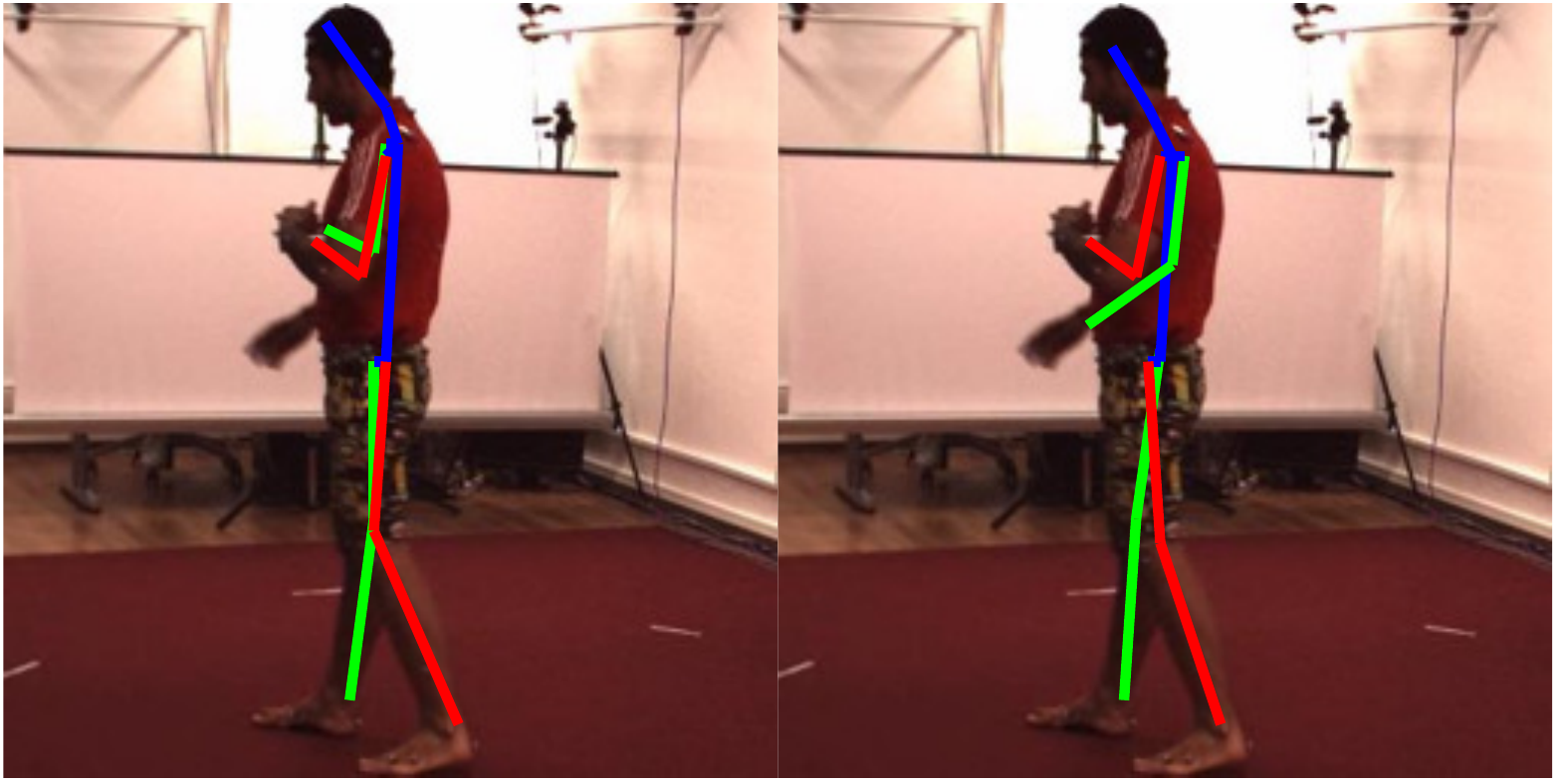}
\includegraphics[width=0.33\textwidth]{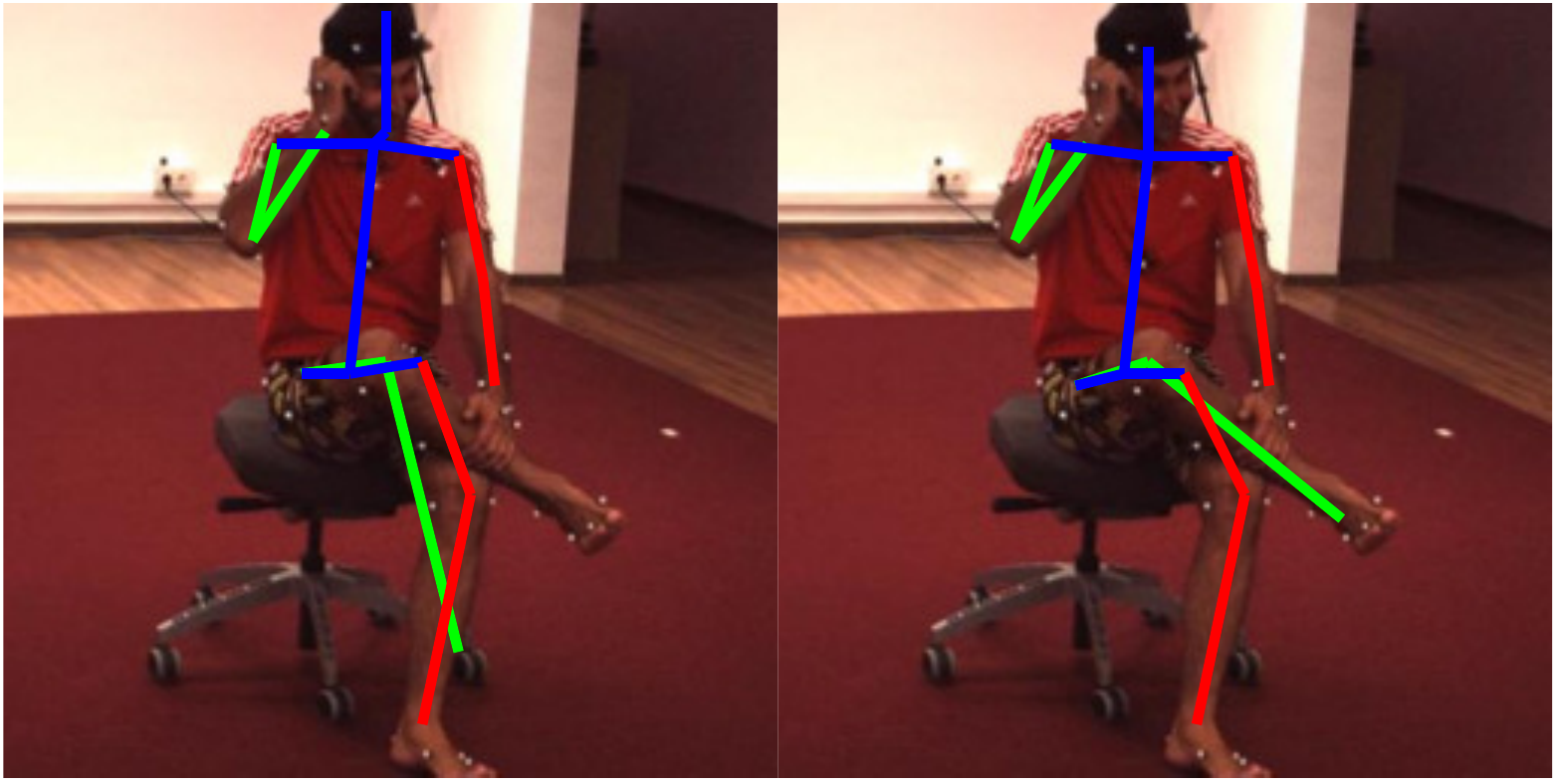} \\
\includegraphics[width=0.33\textwidth]{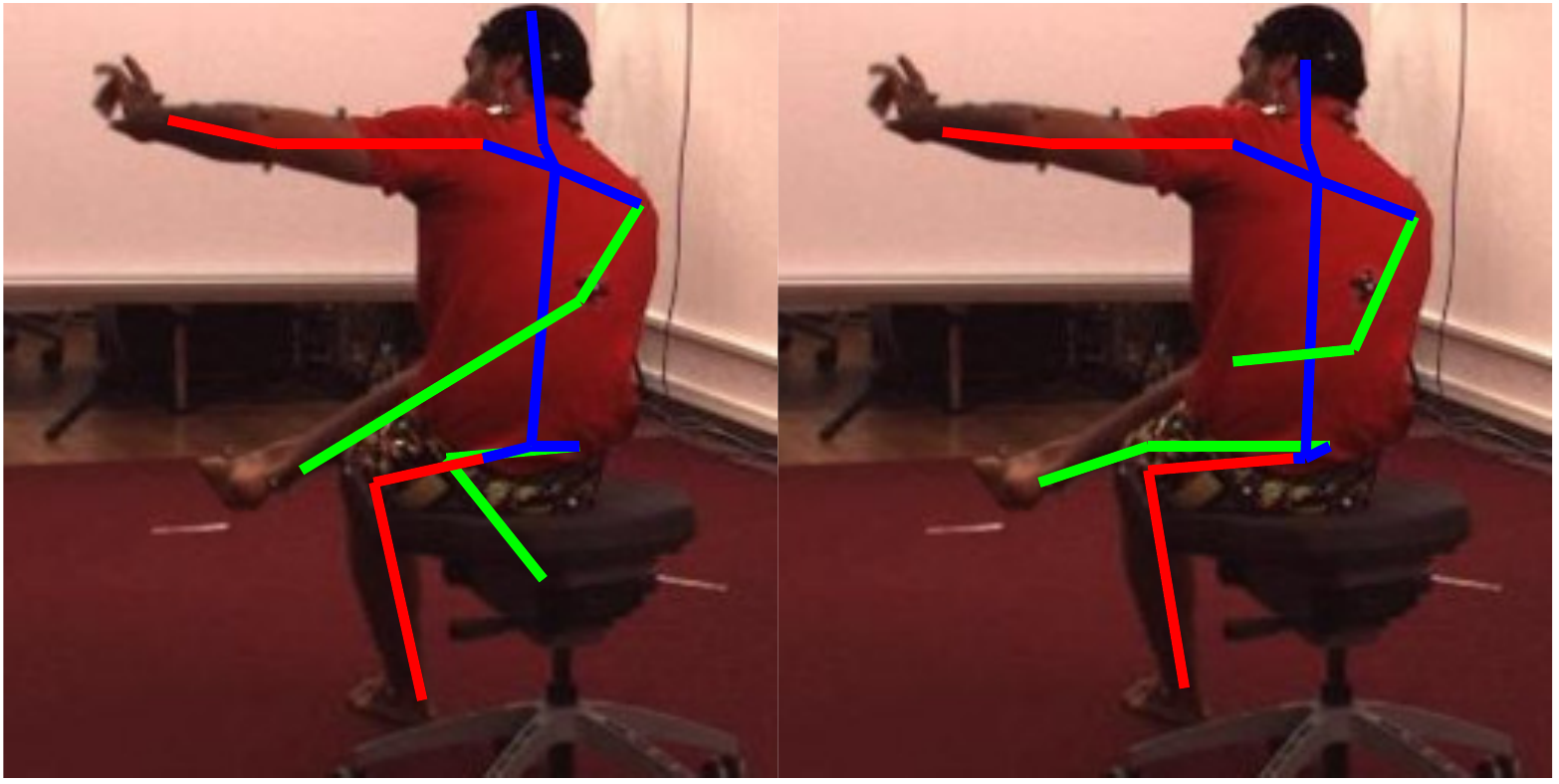}
\includegraphics[width=0.33\textwidth]{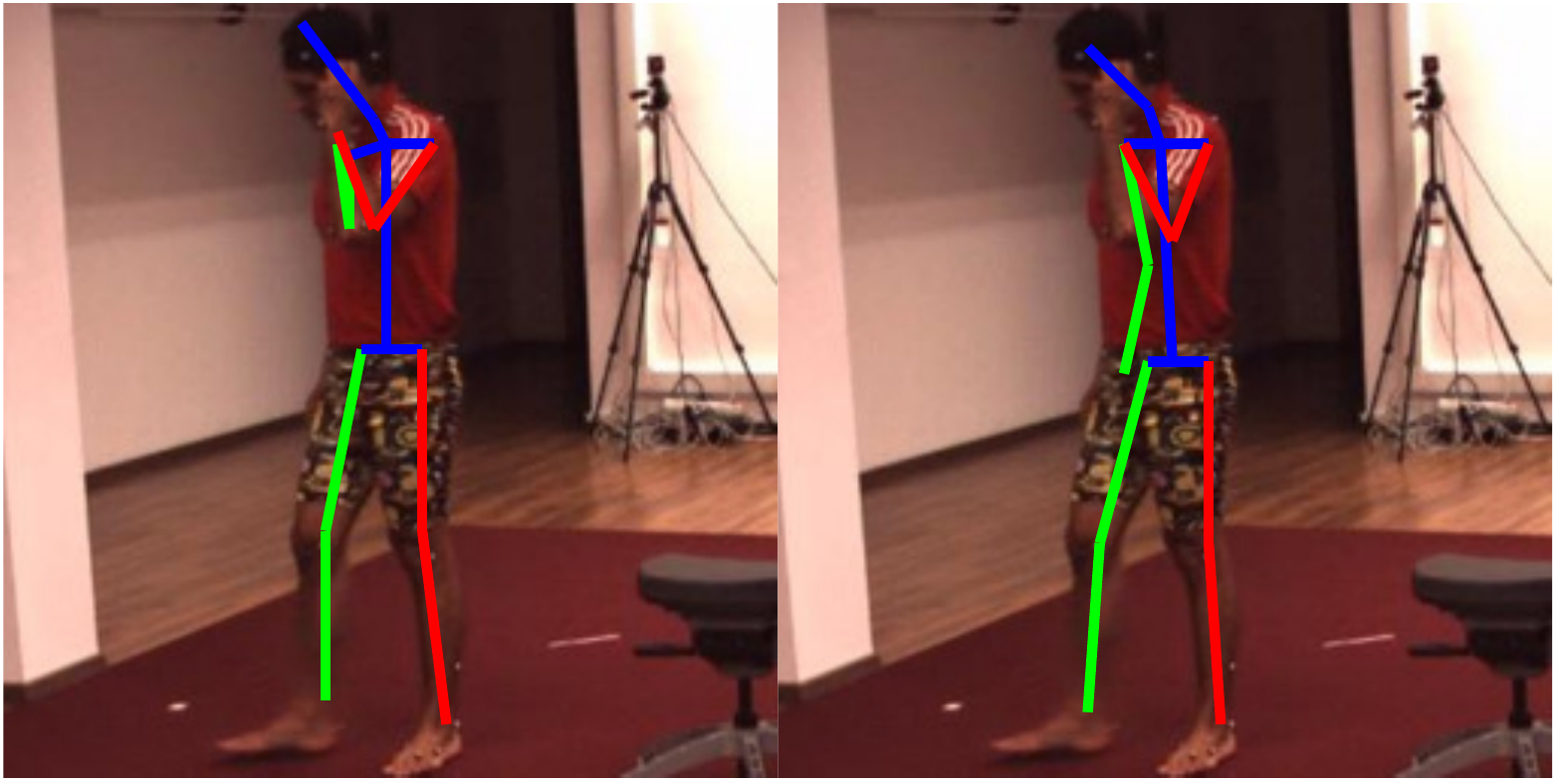}
\includegraphics[width=0.33\textwidth]{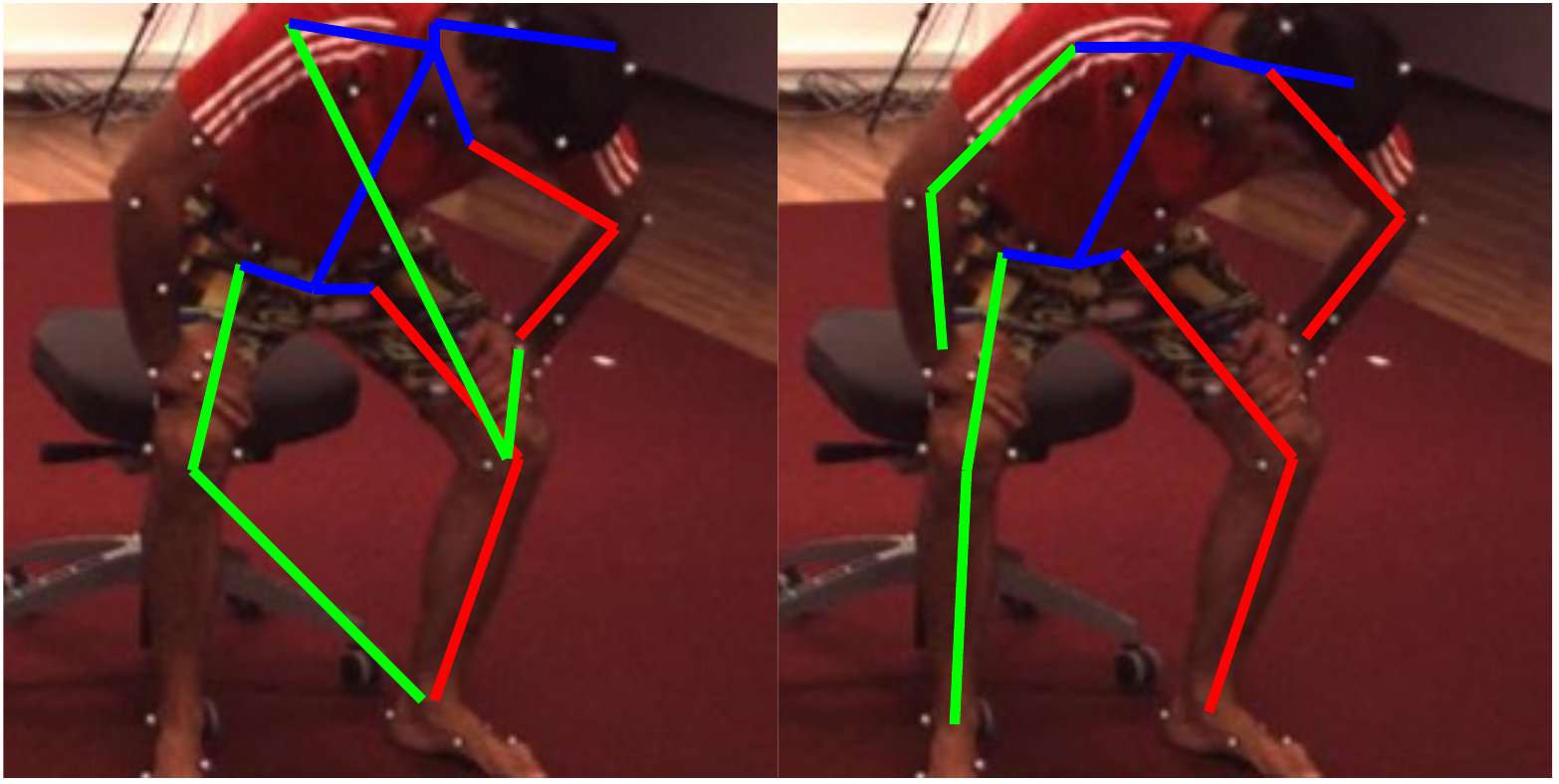} \\
\vspace{-8pt}
\caption{Examples on Human3.6M showing the performance gain from ``personalization''. 
For each pair of images, pose estimation results are presented from the generic (left) and the ``personalized'' ConvNet (right).}
\label{fig:Human3.6_qual}
\end{figure*}

For ``personalization'', Figures  \ref{fig:KTH_qual} and \ref{fig:Human3.6_qual} show qualitative sample results of the proposed approach with and without fine-tuning on annotations recovered from the input imagery on KTH Multiview Football II and Human3.6M, respectively. Despite the generic ConvNet being quite reliable, it might fail for the most challenging poses which are underrepresented in the original generic training set. The benefit from the ``personalized'' ConvNet is greater in these cases since it adapts to the discriminative appearance of the user and recovers the pose successfully.

\begin{figure*}[t]
\includegraphics[width=0.48\textwidth]{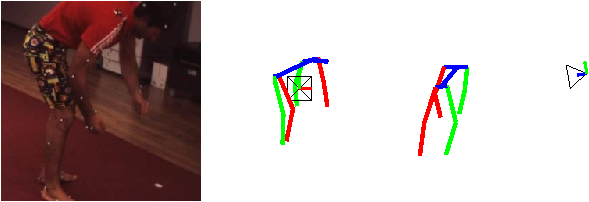}
\includegraphics[width=0.48\textwidth]{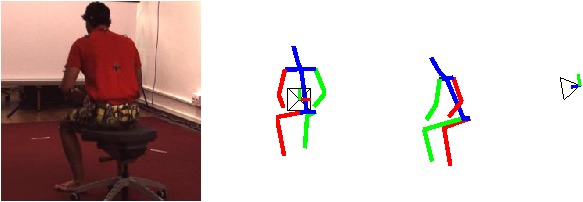} \\ \hspace{0.01em}
\includegraphics[width=0.48\textwidth]{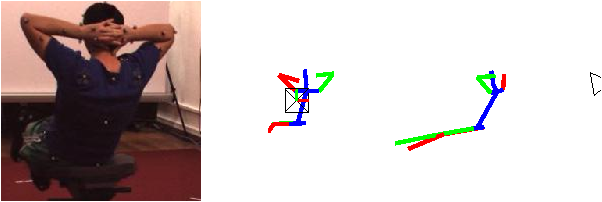}
\includegraphics[width=0.48\textwidth]{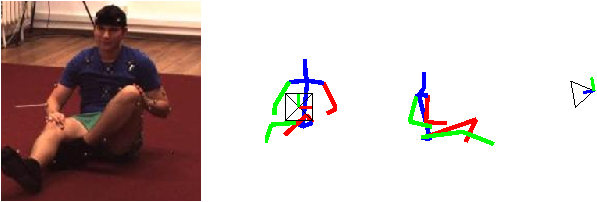}
\caption{Example predictions on Human3.6M from the ConvNet trained to estimate 3D pose from a single image. For each example, we present (left-to-right) the input image, the predicted 3D pose from the original view, and a novel view. Red and green indicate left and right, respectively.
}
\label{fig:HM36M_3d}
\end{figure*}

For the 3D pose ConvNet trained from scratch, we present example 3D reconstructions in Figure~\ref{fig:HM36M_3d}. Notice the challenging poses of the subject and the very accurate poses predicted by the ConvNet.

\section{Summary}

This paper presented an automatic way to gather 3D annotations for human pose estimation tasks, using a generic ConvNet for 2D pose estimation and recordings from a multi-view setup. The automatically generated annotations were used to adapt a generic ConvNet to the particular task, demonstrating important performance benefits from this ``personalization''. Additionally, we trained a ConvNet for 3D pose estimation which performs on par with the current state-of-the-art, even though we only used automatically harvested annotations, and ignored the provided groundtruth.

One promising direction for future work is using the automatic annotation setup in an outdoor environment, (where MoCap systems and depth sensors are not applicable) to collect 3D annotations for in-the-wild images. This would allow us to train a generic 3D human pose ConvNet, similar to the 2D counterparts, by overcoming the bottleneck of limited color images with 3D groundtruth.

\vspace{1em}
\footnotesize
\noindent
{\bf Project Page:} \url{https://www.seas.upenn.edu/~pavlakos/projects/harvesting}

\vspace{0.5em}
\footnotesize
\noindent
{\bf Acknowledgements:} We gratefully appreciate support through the following grants: NSF-DGE-0966142 (IGERT), NSF-IIP-1439681 (I/UCRC), NSF-IIS-1426840, ARL MAST-CTA W911NF-08-2-0004, ARL RCTA W911NF-10-2-0016, ONR N00014-17-1-2093, an ONR STTR (Robotics Research), NSERC Discovery, and the DARPA FLA program.

\clearpage
\clearpage

{\small
\bibliographystyle{ieee}
\bibliography{bibref_definitions_short,bibref}
}
\end{document}